\documentstyle[icip,epsfig]{article}

\title{A Linear Shift Invariant Multiscale Transform}

\name{Andreas Siebert}
\address{Department of Computer Science \\
	 The University of British Columbia \\
	 siebert@cs.ubc.ca }

\begin{document}

\maketitle

\begin{abstract}
{\em
This paper presents a multiscale decomposition algorithm.
Unlike standard wavelet transforms, the proposed operator 
is both linear and shift invariant. 
The central idea is to obtain shift invariance by averaging
the aligned wavelet transform projections over
all circular shifts of the signal.
It is shown how the same transform can be obtained by a
linear filter bank.
}
\end{abstract}

\section{Introduction}
Images typically contain information at many different scales.
For applications such as image retrieval, it is therefore highly desirable
to decompose images orthogonally into subbands. This can be done
using wavelet transforms~\cite{mal89}.

However, the standard wavelet transforms are shift variant,
i.e.~a small shift of the signal can change the wavelet transform
coefficients quite dramatically. As a result, those coefficients
cannot be used for image retrieval when shifts of the original image
can occur.

Several schemes have been proposed recently to introduce shift invariance
into the framework of wavelet transforms.
An early approach was to look at the extrema only of the wavelet
transform modulus~\cite{mh92}.
Several authors suggested the introduction of a cost function in order
to come up with a unique {\em best basis}~\cite{crm95,dhw95,lp94}.
A cost function that figures prominently is the entropy cost function
used in~\cite{cw92}.
Note that the best basis of a structure might change if it is 
embedded into a larger structure.
Filter oriented approaches have been introduced
in~\cite{hkn96} and~\cite{zfg94}.

Most approaches cited above use a notion of shift invariance,
called $\mu-translation-invariance$ in~\cite{lp94}
and {\em near\ shift\ invariance} in~\cite{hkn96},
that is different from the standard usage.
They take shift invariance to imply that the wavelet transform coefficients
do not change at all if the signal is shifted.
However, they extract the different {\em delays} of translated signals.
By contrast, we stick with the standard definition of shift invariance
which means that the operator output is shifted proportionally
to the shift of the input signal.
In~\cite{sfah92}, it is shown that shift invariance (referred to as
{\em shiftability} in that paper) cannot be achieved by
a critically sampling transform like the wavelet transform
because of aliasing. 

In this paper, we propose a decomposition algorithm called the
{\em Averaged Wavelet Transform}~(AWT) that incorporates
shift invariance in the standard sense.
It delivers a unique representation of a signal without
requiring a cost function.

Systems using wavelets for the purpose of classification and
retrieval have been described in~\cite{jfs95,opsop97,tsdgkdw94}.

\section{The Averaged Wavelet Transform}
In one dimension, we define the AWT to be
a mapping from a discrete signal $SIG$ of length $n$ to $k+1$ 
vectors of the same length $n$, where $k$ is the number of
decomposition levels which is logarithmic in $n$. That is,
\[
AWT: R^n \rightarrow R^n \times R^n \times \dots \ \ (\mathrm{k+1\ \ times})
\]

The key idea for obtaining a shift invariant signal decomposition
is to average the {\em projections of the signal onto the wavelet subspace} 
at each scale over all circular shifts of the input signal.
We call the result the {\em scale spectra} of the signal.
Note that those projections are sometimes also referred to as
{\em detail vectors}, or {\em reconstructed details}, or just
{\em details}.\footnote{In Matlab, the details are reconstructed by calling the function
{\tt wrcoef()}. It is the inverse of the wavelet transform.}

More precisely, let $WT$ be a discrete wavelet transform,
let $CS_i$ be the circular right shift operator (performing $i$ right shifts),
with $CS_i^{-1}$ analogously being the circular left shift operator,
and let $REC_s$ be the reconstruction operator that
projects the signal onto the wavelet subspace
at scale $s, s=0..k$,
with the special case $s_0$ being the DC
component.\footnote{In Matlab, the DC component is referred to as
{\em reconstructed approximation coefficients}.}

Then the AWT at scale $s$ is
\[
AWT_s = \frac{1}{n} \sum_{i=0}^{n-1} CS_i^{-1}(REC_s(WT(CS_i(SIG))))
\]
The left shift operator $CS_i^{-1}$ is applied to compensate for the
shifting of the signal, i.e.~we average over the projections
that are aligned to the position of the signal before shifting.
The complete AWT is the set of all $AWT_s, s=0..k$. 

The shift invariance of this approach follows directly from the
nature of the averaging process: if the input signal is shifted,
the AWT is still computed over the same set of $n$ circularly shifted signals.
Also, as an inherent property of the averaging operation, the AWT is linear.
Like the standard WT, the AWT decomposes a signal into a DC component and
scale spectra of zero mean, with some frequency overlap 
between the scale spectra.
Adding a constant to all signal values
only changes the DC component, but not the scale spectra.

The AWT is information preserving.
The original signal can be easily reconstructed
by just adding up the scale spectra, i.e.
\[
SIG = \sum_{s=0}^k AWT_s
\]
In this sense, the summation is the inverse transform of the AWT.
Note that the AWT is highly redundant.
It increases the space requirement by a factor of $k+1$.
However, for most applications, the scale spectra will serve
only as an intermediate representation.
Typically, it suffices to keep only a few features
which are derived from the scale spectra.

The generalization from 1-d to 2-d is conceptually straightforward, albeit
computationally expensive.
The circular shifting has to be done over all shifts along the two axes,
and the 1-d wavelet transform is replaced by a 2-d WT.
Fortunately, an efficient algorithm to compute
the wavelet coefficients of all $n$
circular shifts of a vector of length $n$ in $O(n\ \mathrm{lg}\: n)$ time,
rather than the $O(n^2)$ time of the naive approach, is given in~\cite{bey92},
based on the observation that the shift operation is sparse.
\cite{lp94} describes the extension of Beylkin's algorithm to 2-d.

\section{Emulation of the AWT by a Filter Bank}

As pointed out above, the AWT is a shift invariant linear transform.
As such, the AWT is well defined by a single impulse response
for each scale spectrum.
This means that we have to compute the underlying set of 
{\em AWT filters} only once,
and then the AWT can be computed simply and efficiently by convolution
of the signal with those AWT filters.

The filter bank that emulates an AWT depends on the type of wavelet
chosen for the wavelet transform and on the size of the signal -- each
signal size has its own corresponding AWT filter bank.
The AWT filters are symmetric and therefore linear phase.
By construction, their length is
bounded by the signal length, so they are FIR filters. 

\begin{figure}[ht]
  \epsfig{file=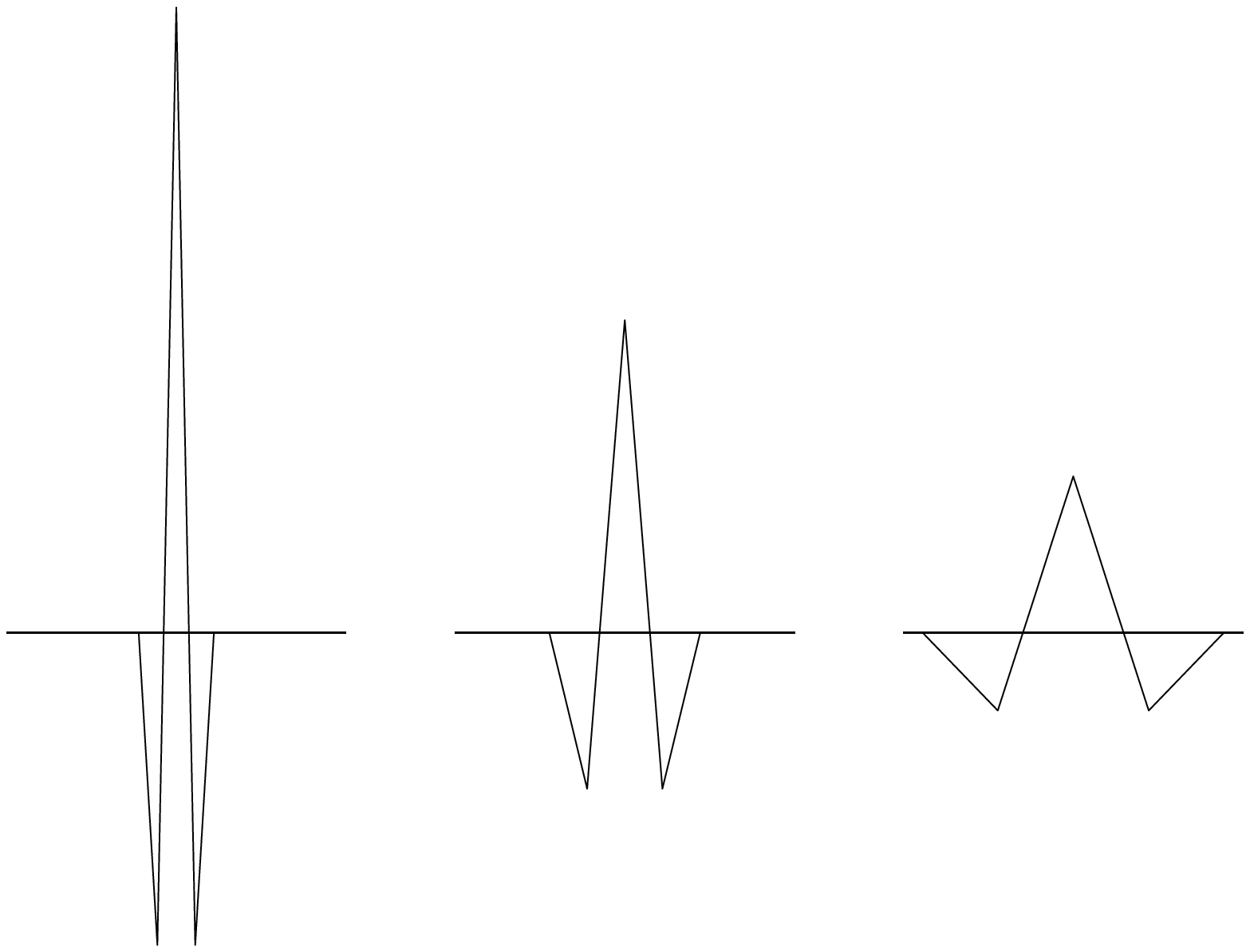,width=8.3cm, height=2.0cm}
  \epsfig{file=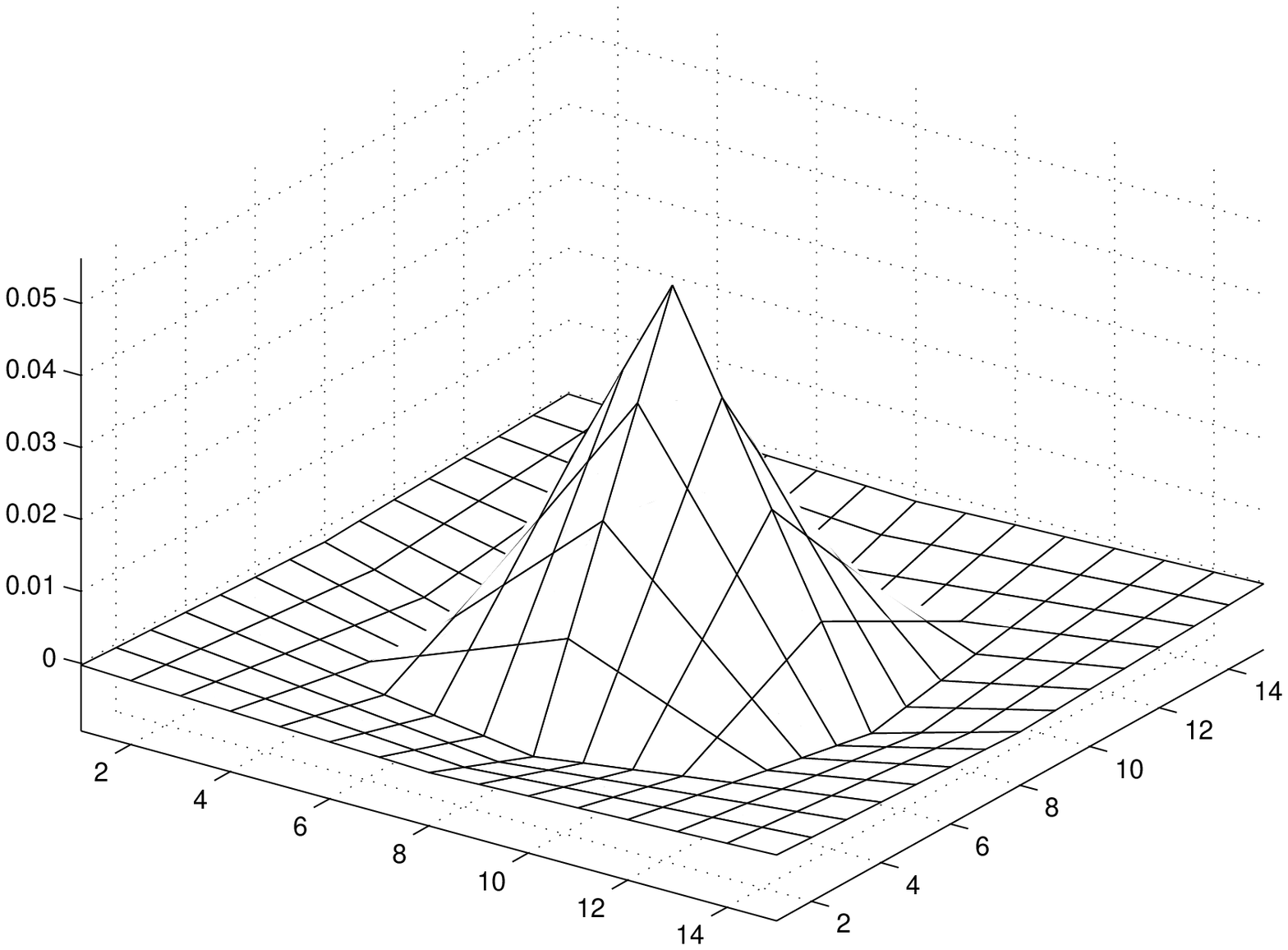,width=4.1cm}
  \epsfig{file=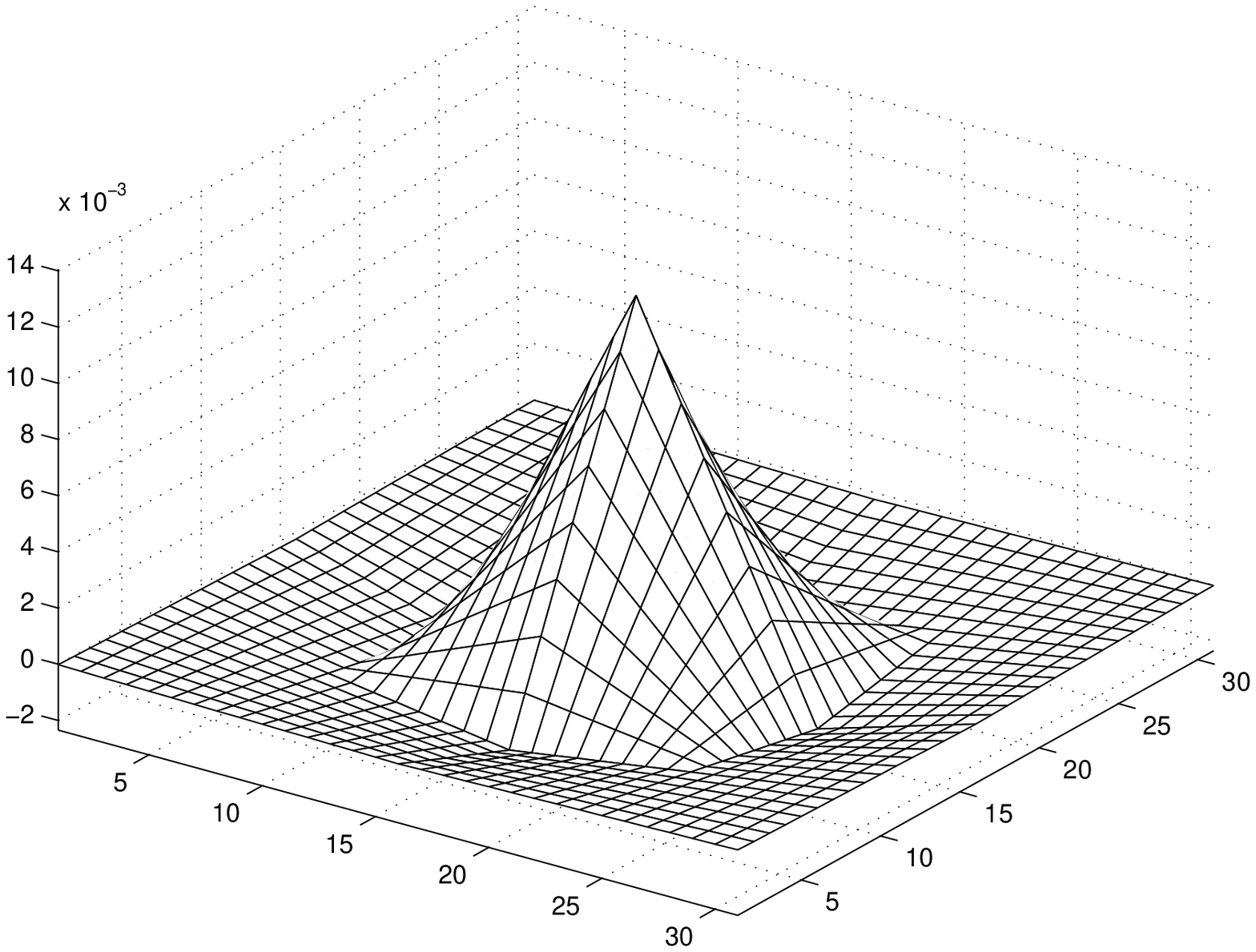,width=4.1cm}
  \caption{AWT filters (Haar), 1-d and 2-d, consecutive scales,
	signal length is a power of two}
  \label{fig:HAARFilt}
\end{figure}

\begin{figure}[ht]
  \epsfig{file=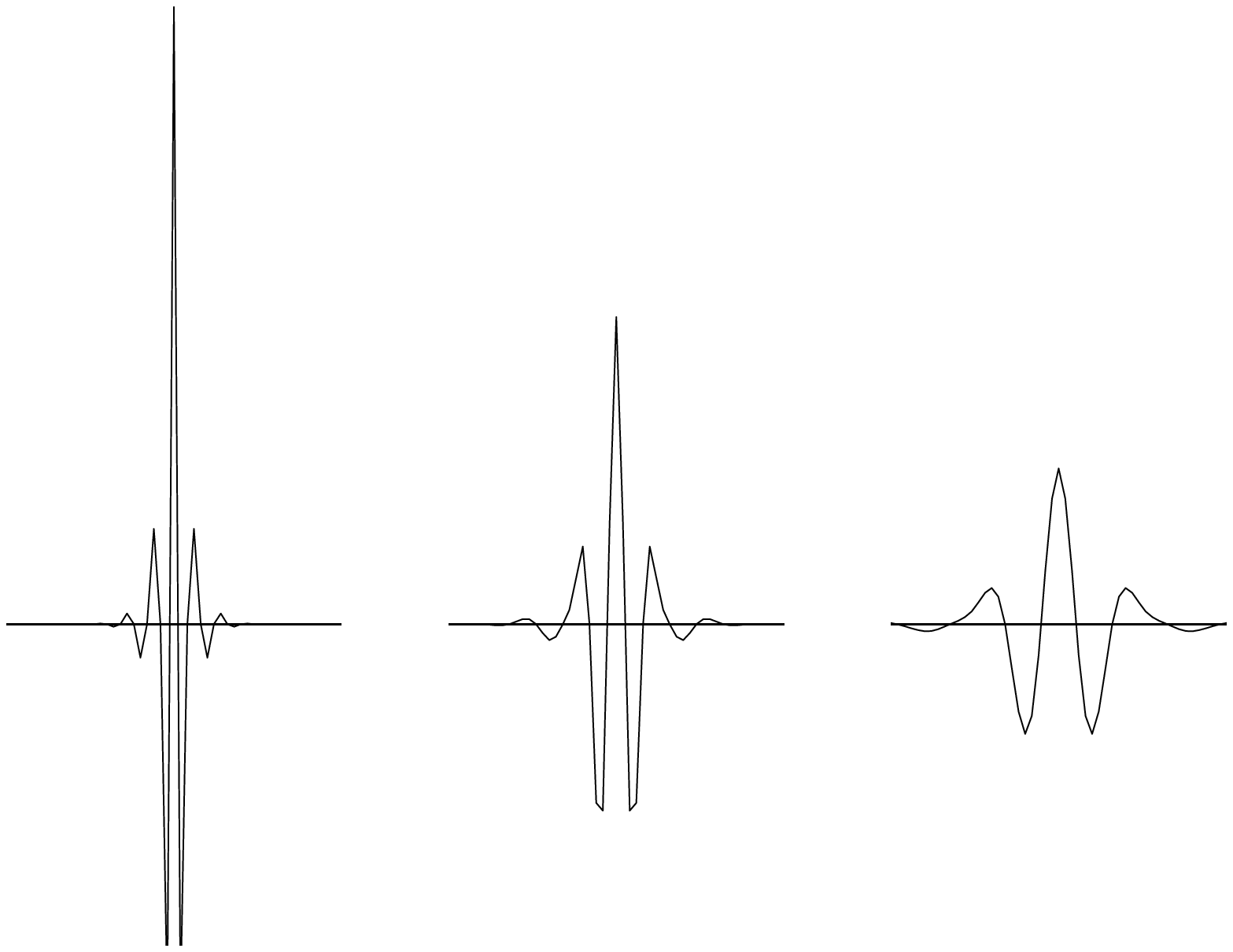,width=8.3cm, height=2.0cm}
  \epsfig{file=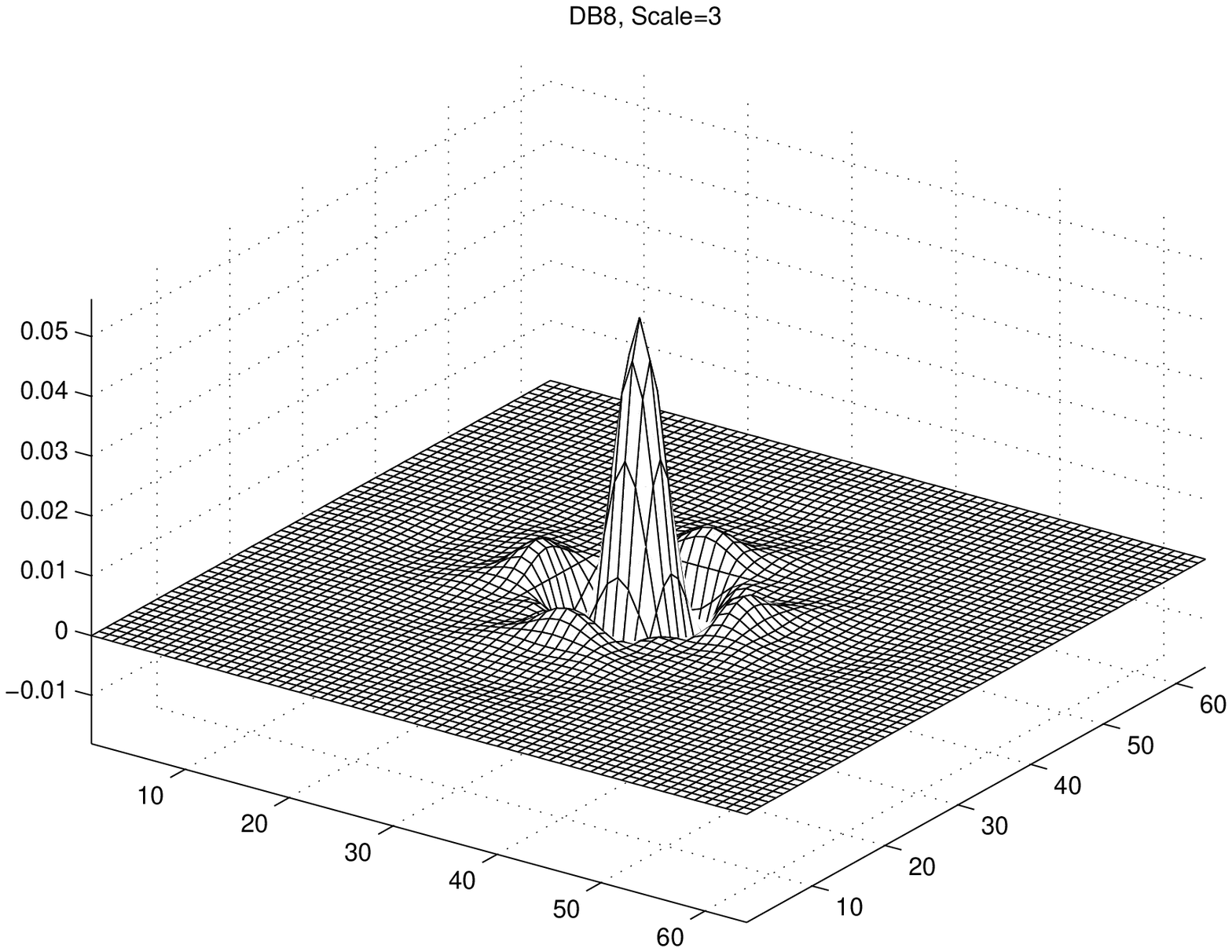,width=4.1cm}
  \epsfig{file=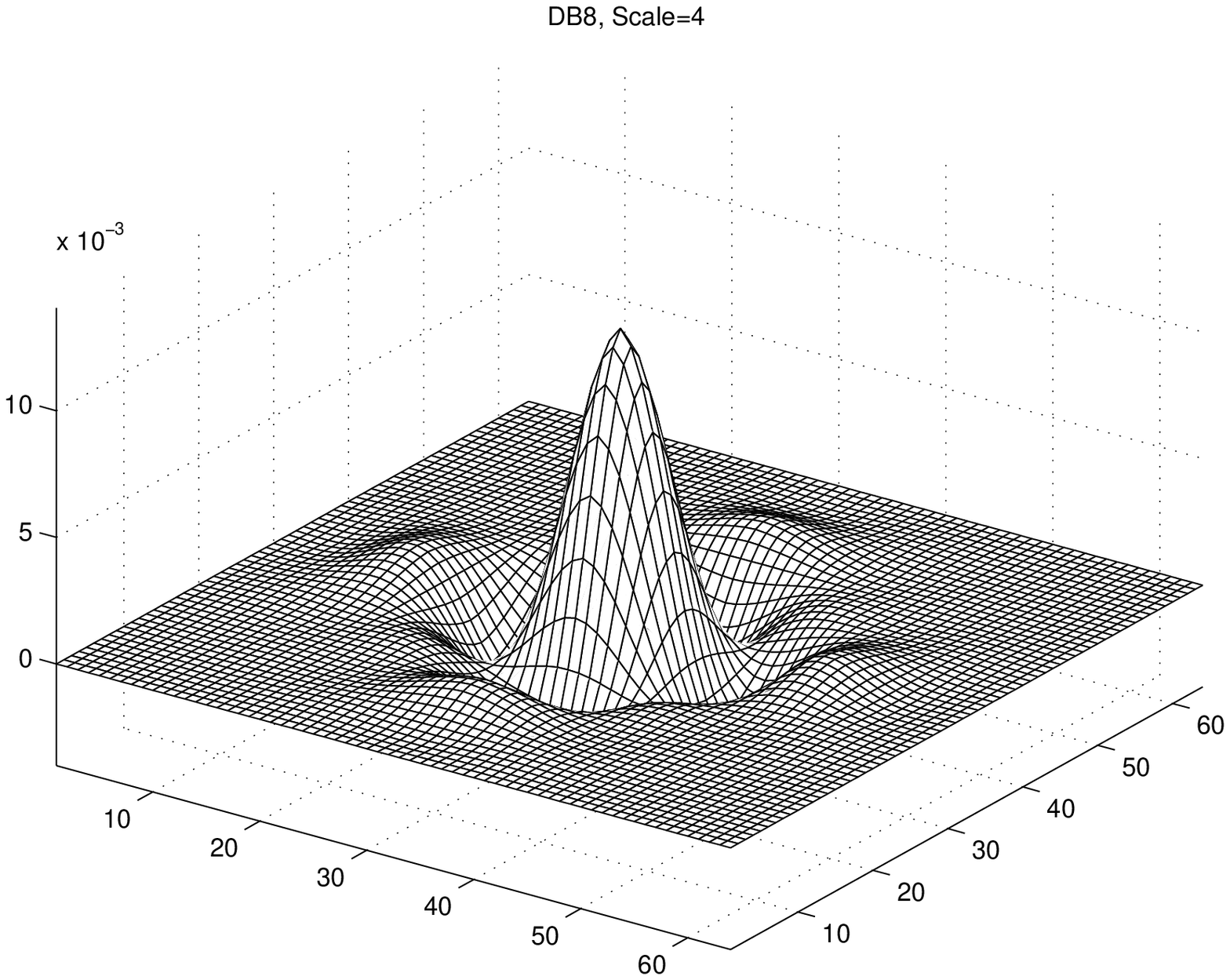,width=4.1cm}
  \caption{AWT filters (Daub$_8$), 1-d and 2-d, consecutive scales,
	signal length is a power of two}
  \label{fig:DAUBFilt}
\end{figure}

The simplest AWT filter bank is obtained for the Haar wavelet with a signal
length that is a power of two. They turn out to be dilations of the
hat function, shown in fig.~\ref{fig:HAARFilt} for the first three 1-d
scales and for two scales in the 2-d case.
In the 1-d case, the filter coefficients are
{\tt (-0.25 0.5 -0.25), (-0.0625 -0.125 0.0625 0.25 0.0625 -0.125 -0.0625)}
for scale one and two, respectively. 
The AWT filters $y^{(i)}(x)$ at scale~$i$
are easily computed from the filter
at scale~$1$ by the formula $y^{(i)}(x)=2^{-(i-1)}y^{(1)}(2^{-(i-1)}x)$
which nicely reflects their wavelet transform origin.
Although these AWT filters are the solution of a dilation equation,
they are {\em not} wavelets~\cite{str89}: they are orthogonal
to only some of their translates and
dilations.\footnote{\cite{str89} says on p.619
``It [the wavelet W(x) from the hat fct.]
is not orthogonal to W(x+1).'' What was meant was that the
{\em scaling function} is not orthogonal to its translates
[personal communication with the author].}
This comes as no surprise, since it was shown in~\cite{sfah92}
that shiftability can only be achieved by relaxing
the orthogonality property of the wavelet transform.
Clearly, the AWT is not a pure wavelet transform, and it doesn't
provide for a multiresolution~\cite{mal89}.
Note also that the formula for AWT filters for Haar wavelets
does not generalize
to signal sizes that are not a power of two or to other types
of wavelets.  
Fig.~\ref{fig:DAUBFilt} shows some of the AWT filters derived from
Daubechies-8 wavelets.

\section{Experiments}
\begin{figure}[ht]
  \epsfig{file=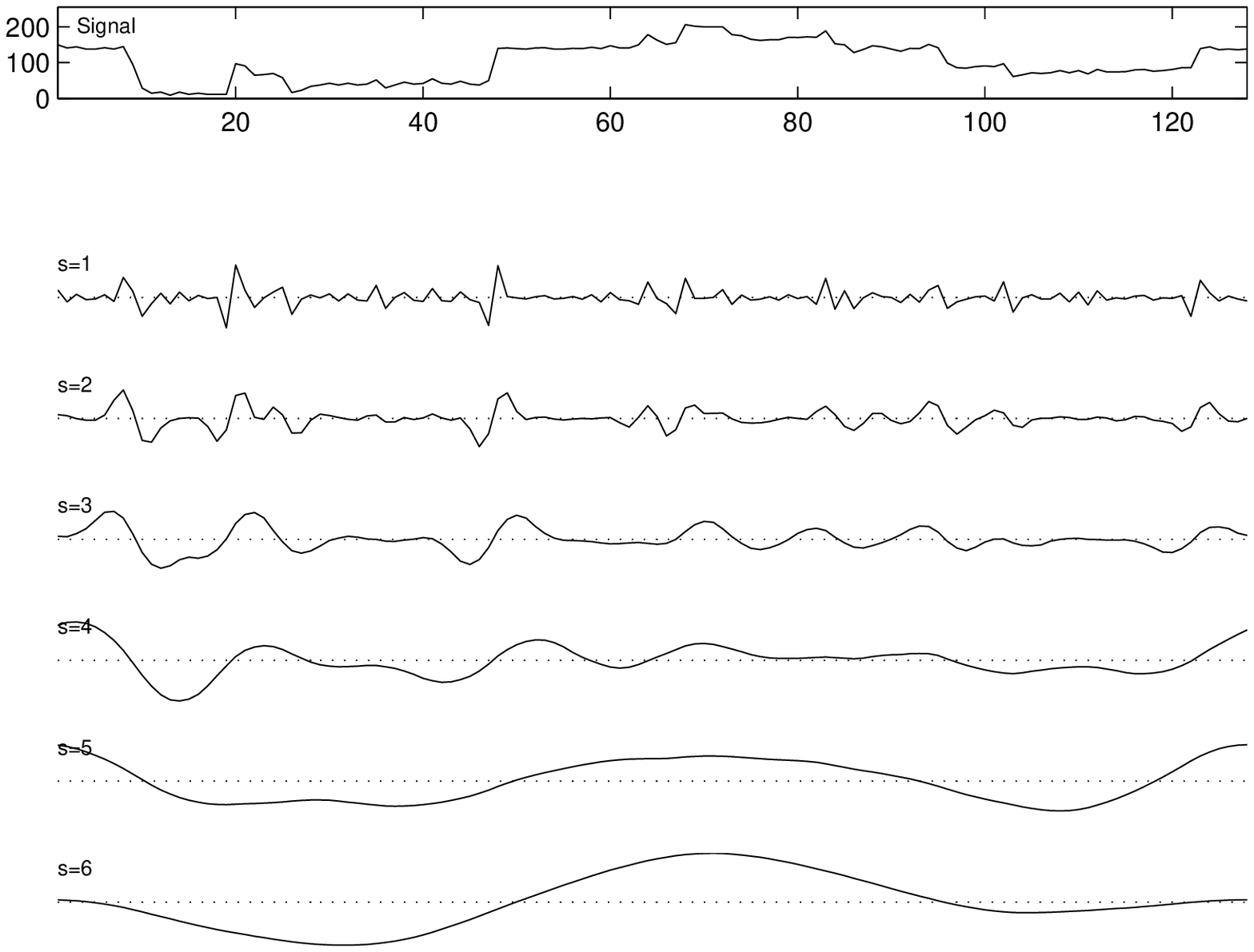,width=8.3cm,height=4.5cm}
  \epsfig{file=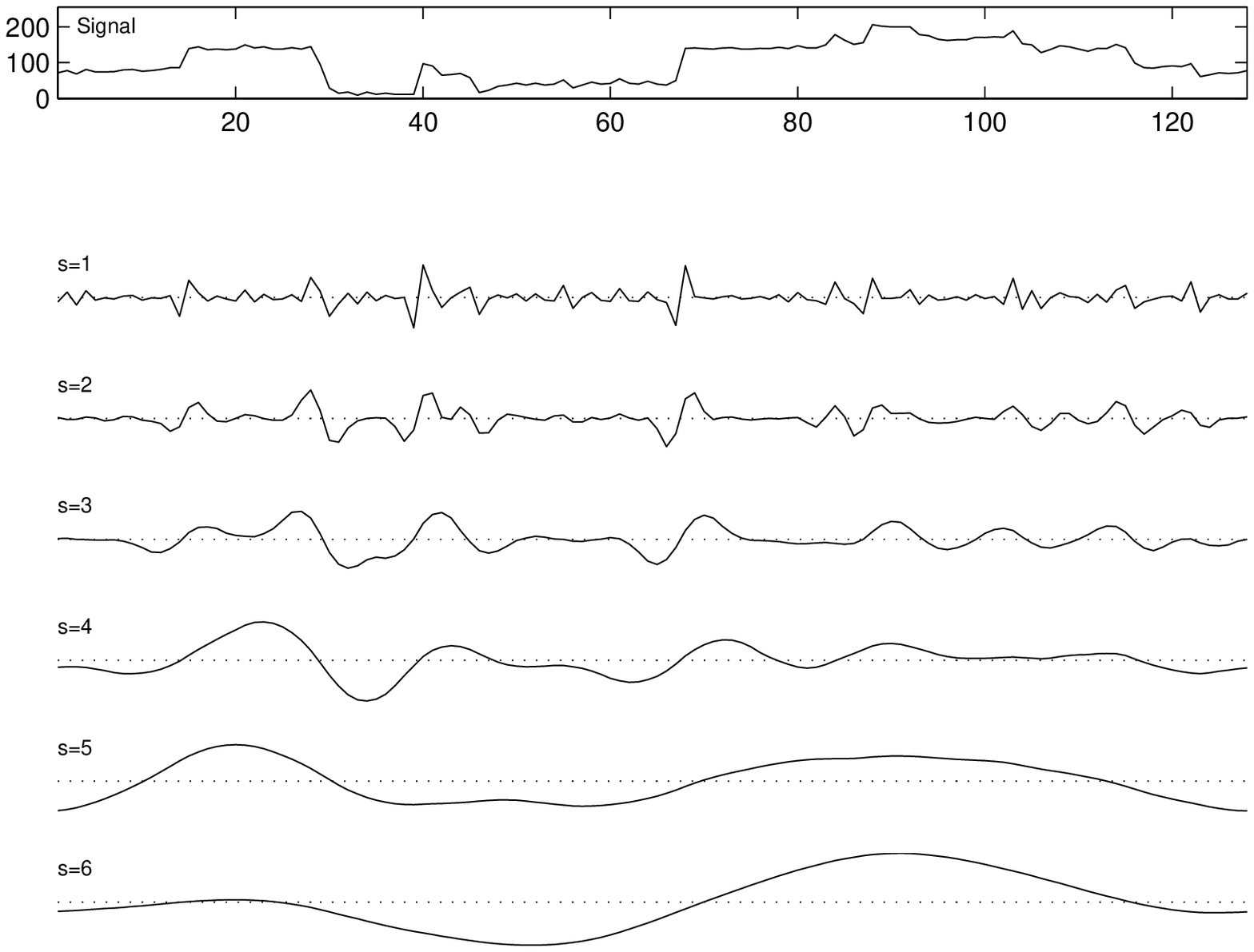,width=8.3cm,height=4.5cm}
  \caption{(top) AWT of a signal,\ \ (bottom) AWT of the same signal,
	but circularly right shifted by 20 units.}
  \label{fig:ShiftOri}
\end{figure}

\begin{figure}[ht]
  \epsfig{file=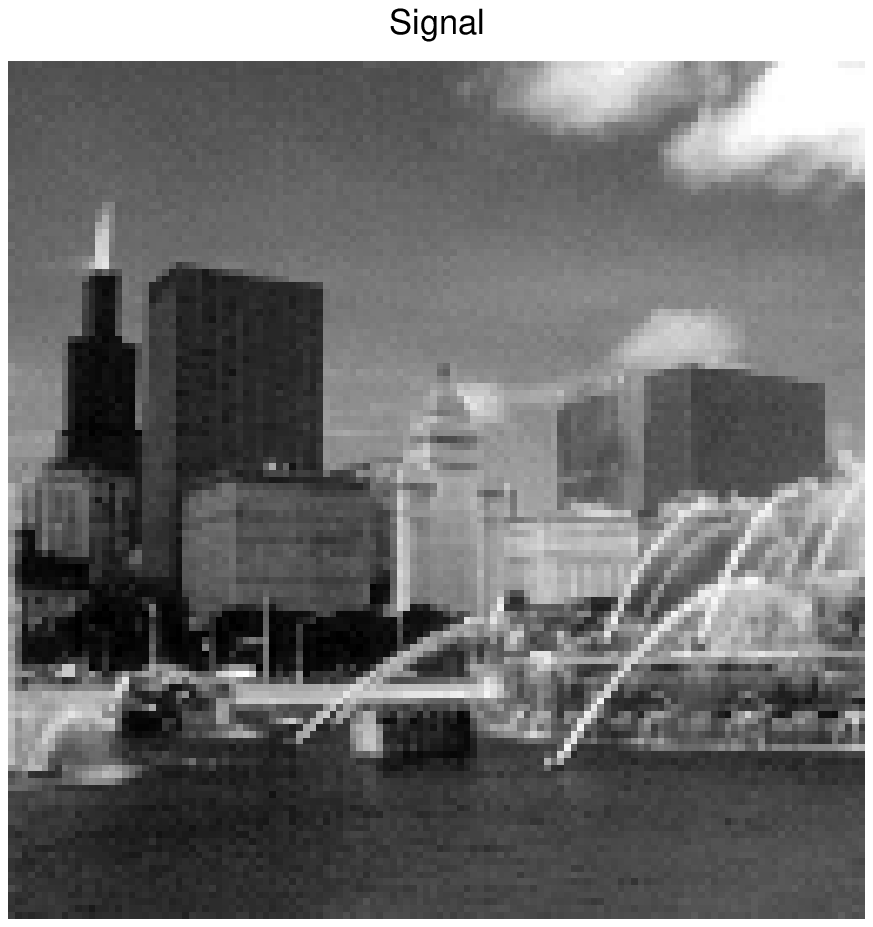,width=4.1cm}
  \epsfig{file=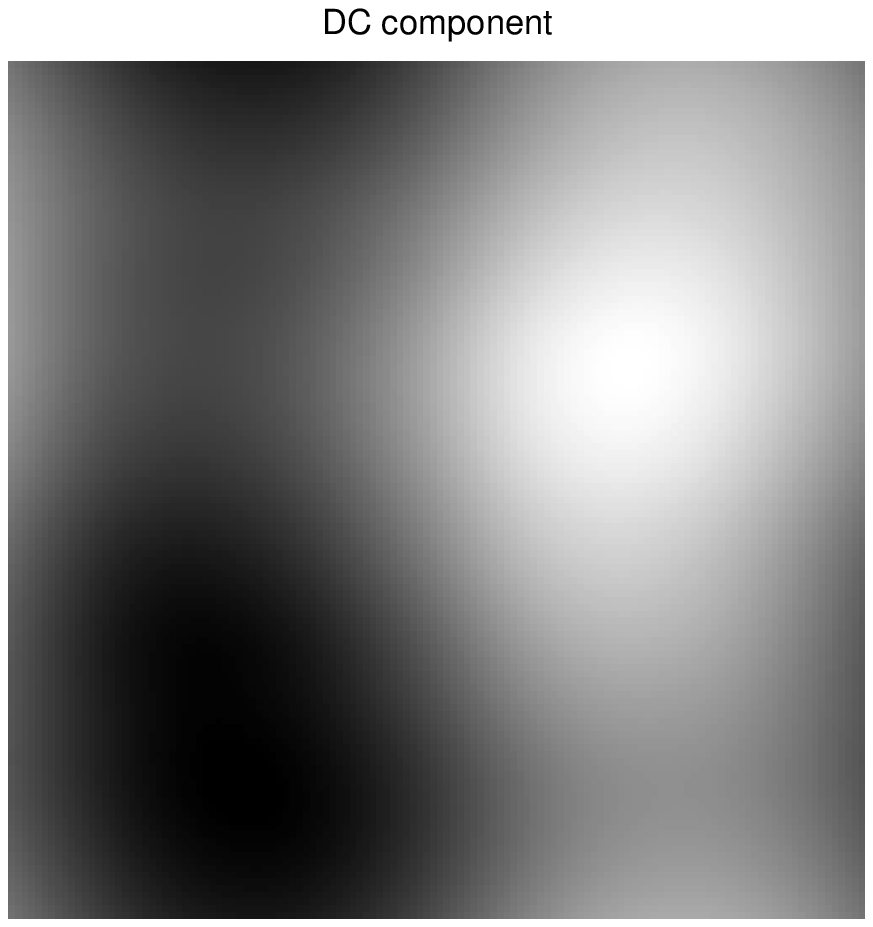,width=4.1cm}
  \epsfig{file=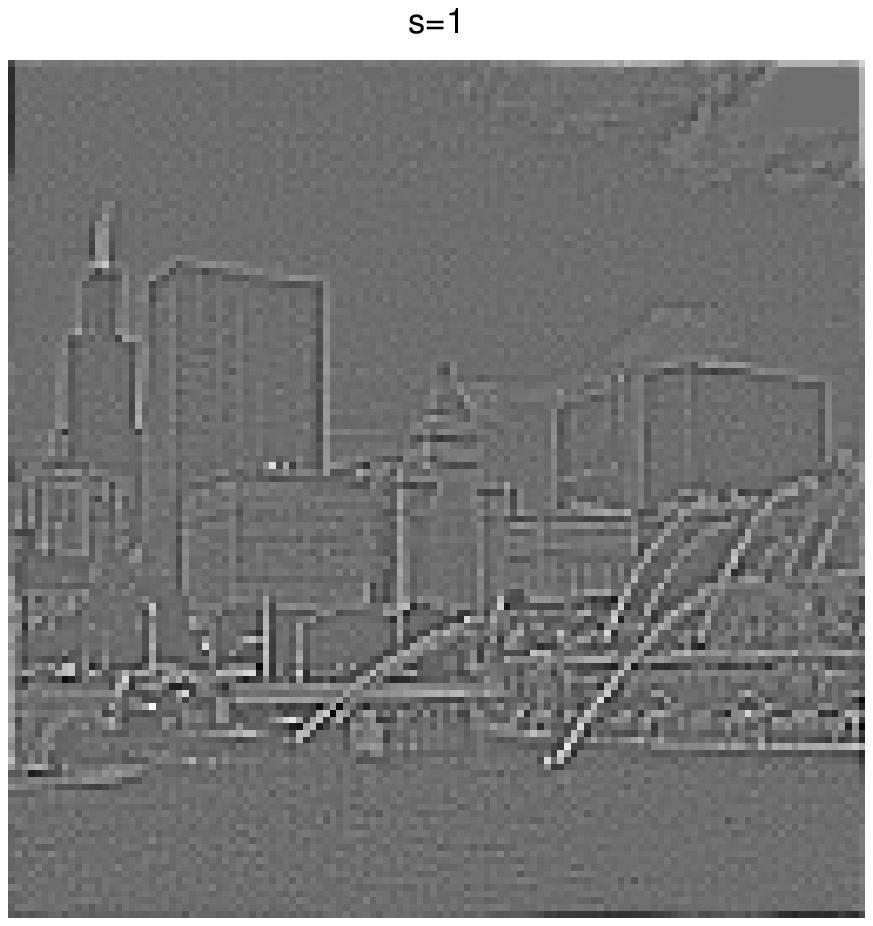,width=4.1cm}
  \epsfig{file=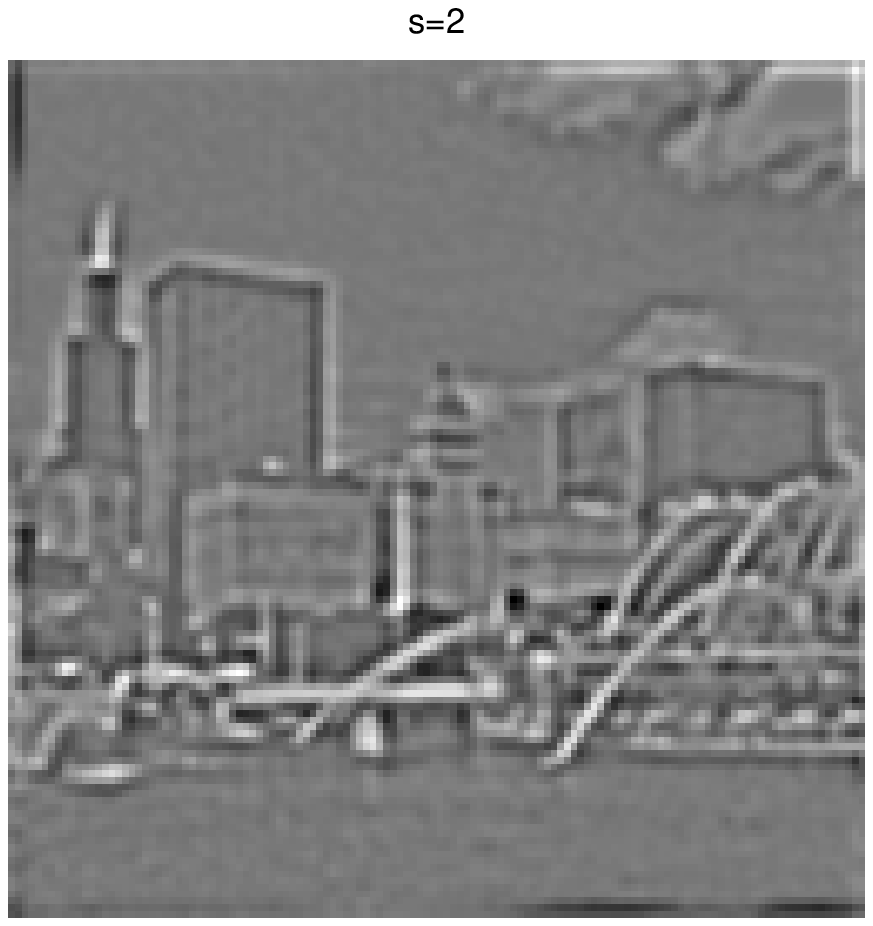,width=4.1cm}
  \epsfig{file=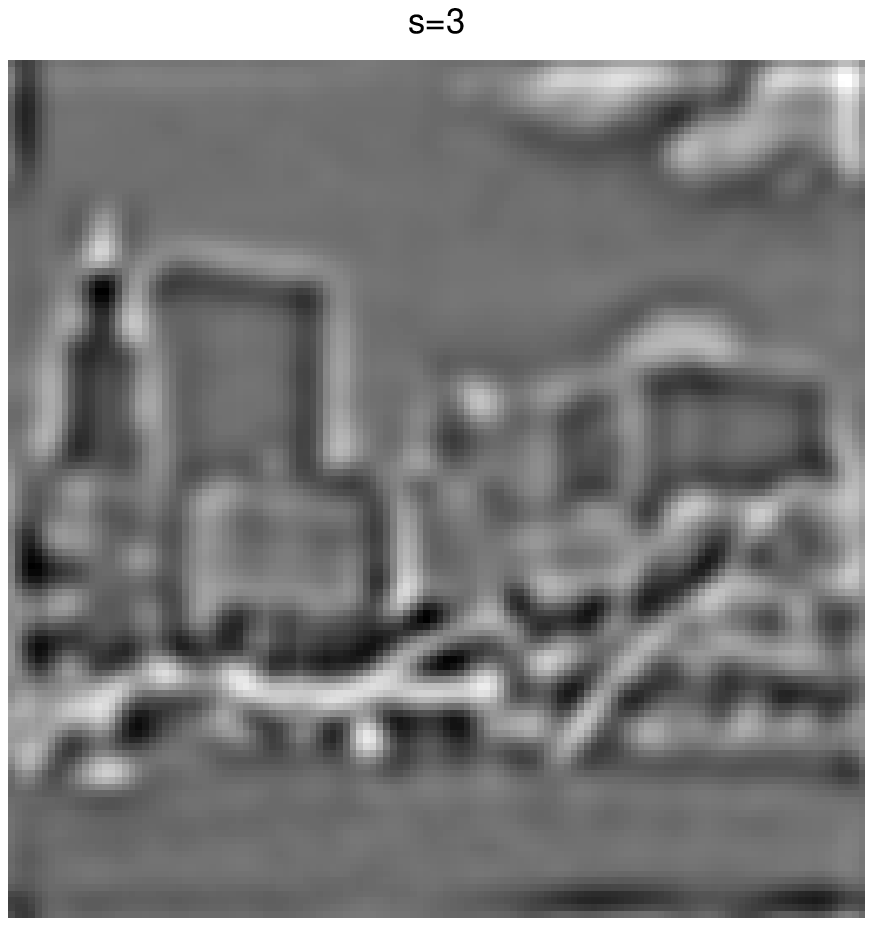,width=4.1cm}
  \epsfig{file=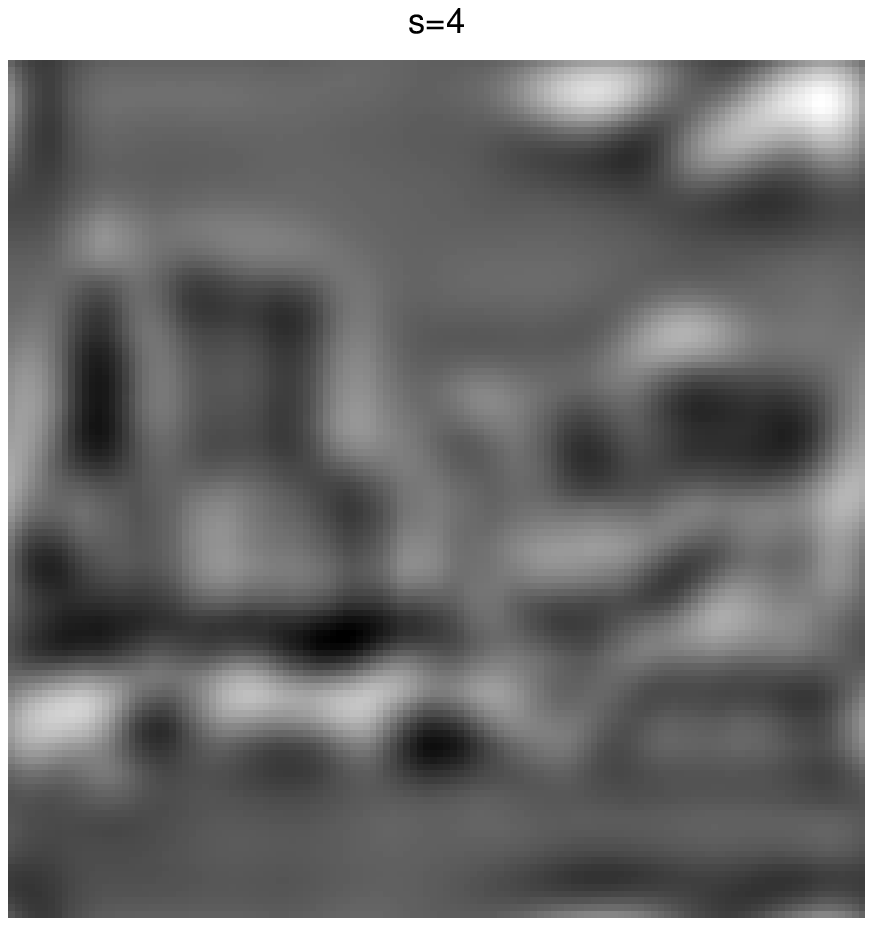,width=4.1cm}
  \epsfig{file=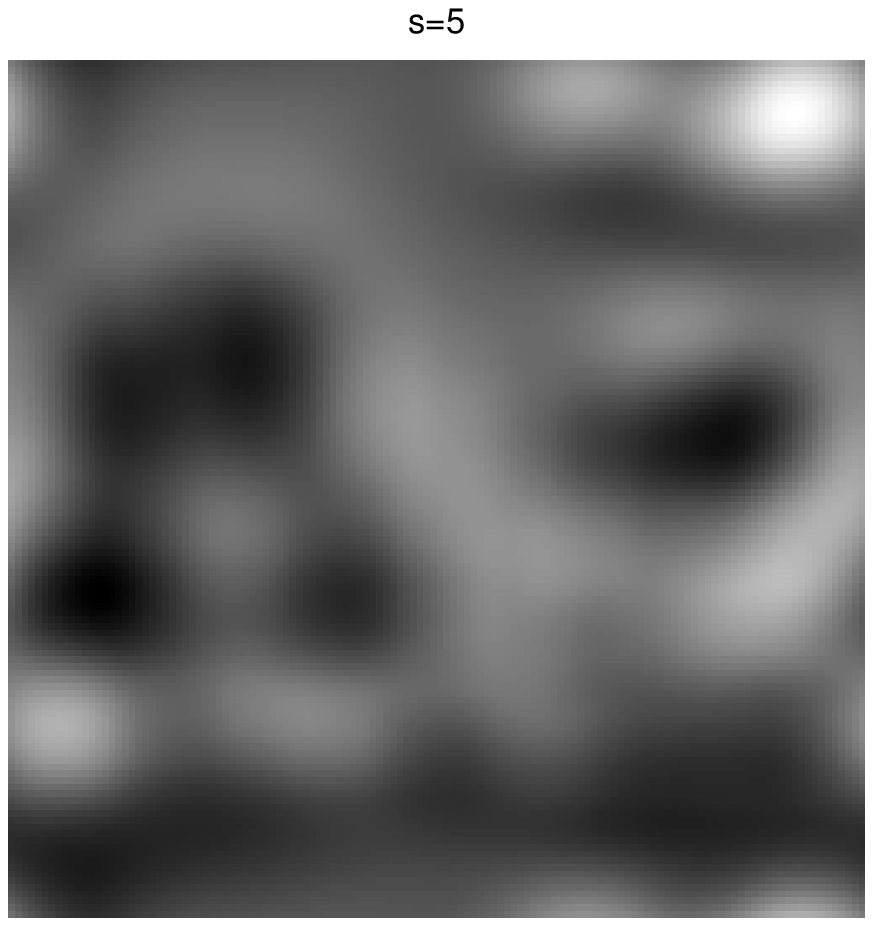,width=4.1cm}
  \epsfig{file=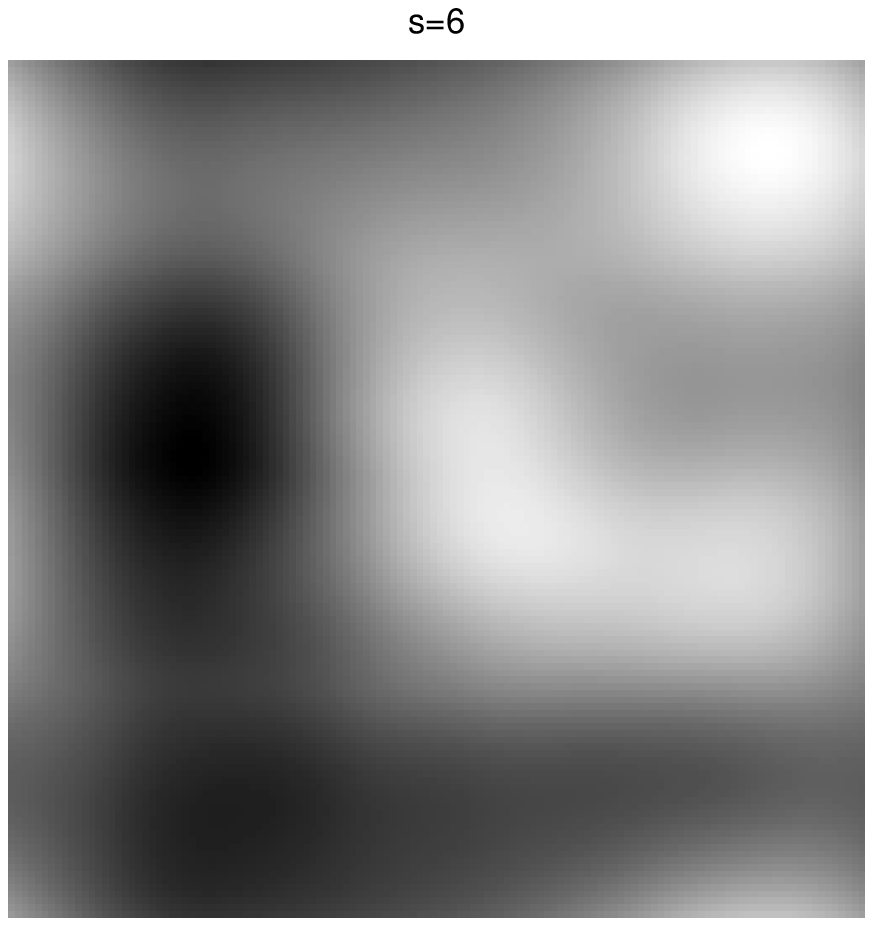,width=4.1cm}
  \caption{2-d AWT (Haar, 128$\times$128).
	Original Image, DC component, six dyadic scales.}
  \label{fig:Chic2d}
\end{figure}

Figure~\ref{fig:ShiftOri} demonstrates the decomposition of 
a 1-d signal (it happens to be a cross section of the image
in fig.~\ref{fig:Chic2d}) by the AWT,
and the shift invariance of the latter: all scale spectra of the
circularly shifted signal are also circularly shifted by the same amount.
We have chosen Haar wavelets here, but the AWT works for any wavelet.
While a standard Haar wavelet transform results in 
projections that are step functions,
the AWT gives a much smoother decomposition.

\begin{figure}[ht]
  \epsfig{file=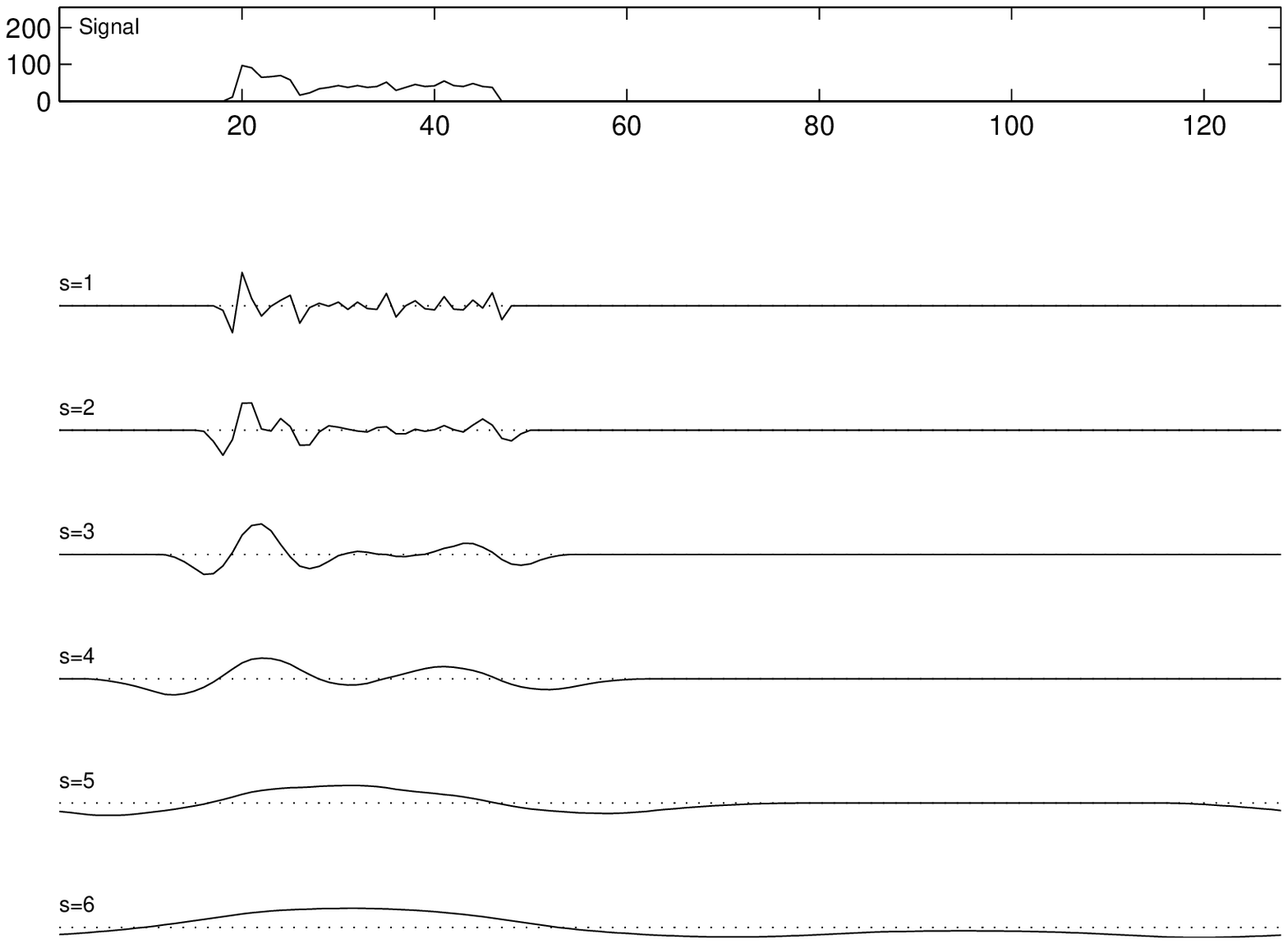,width=8.3cm,height=4.5cm}
  \epsfig{file=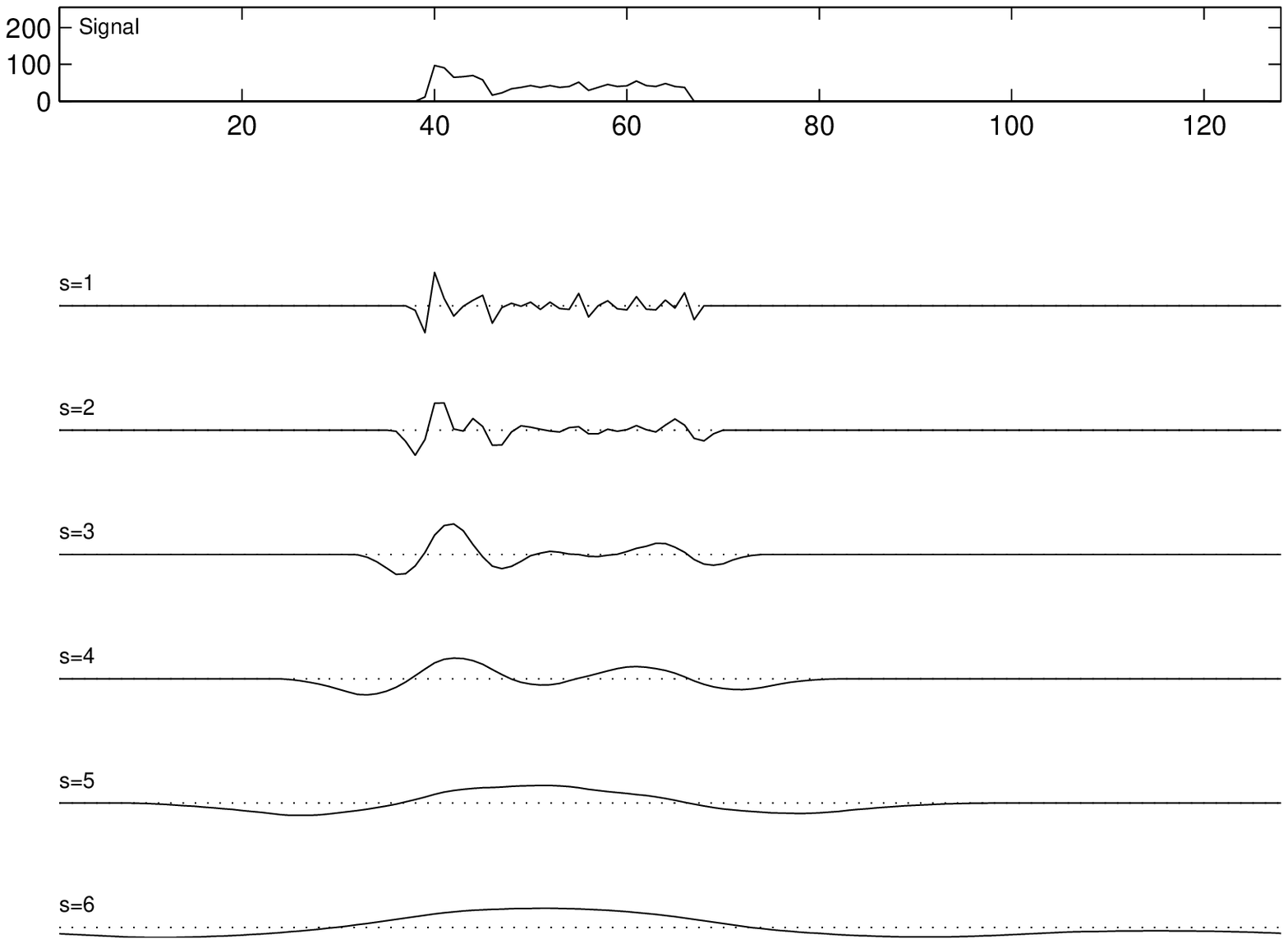,width=8.3cm,height=4.5cm}
  \caption{(top) AWT of a substructure of the signal in~fig.~\ref{fig:ShiftOri},
	\ \ (bottom) AWT
	of the shifted substructure.}
  \label{fig:ShiftSub}
\end{figure}

Figure~\ref{fig:ShiftSub} demonstrates the usefulness of the AWT
for the retrieval of image substructures at various scales.
The signal now consists of a small substructure of the signal
in fig.~\ref{fig:ShiftOri}, with the rest of the signal being set to zero.
Again, the AWT decomposition of this signal
is given at the top, and the AWT
of the same substructure, circularly right shifted, at the bottom.

It can be observed that the AWT of the substructure matches the 
corresponding part of the AWT of the full signal to a large degree,
depending on the scale.
At small scales, the filter support is small and therefore localized.
By contrast, at large scales, the area of support is large,
which means that the filter response to a small substructure
is influenced to a large degree
by the surrounding signal, i.e.~adjacent structures.
It is also evident that the substructure cut-off may introduce 
an edge artifact.

Figure~\ref{fig:Chic2d} gives an example of applying the 2-d~AWT,
based on the Haar wavelet, to an image. The six scale spectra plus
the DC component add up to the original signal.

\clearpage

\subsection*{Acknowledgements}
I would like to thank B.~Woodham for his valuable critique.


\end{document}